%% file: root.tex
\documentclass[letterpaper, 10 pt, conference]{ieeeconf} 

\IEEEoverridecommandlockouts                              

\usepackage[utf8]{inputenc}
\usepackage{epsfig} %
\usepackage{sidecap}

\usepackage{bbold}
\usepackage{mathtools}

\usepackage{floatrow}
\usepackage{mathptmx} 
\usepackage{times} 
\usepackage{amsmath} 
\usepackage{amssymb}  
\usepackage{mathrsfs}
\usepackage{caption}  
\usepackage{subcaption}
\usepackage{algpseudocode}
\usepackage{algorithm}
\usepackage{hyperref}
\usepackage{bbm}
\usepackage{booktabs}
\usepackage{multirow}
\usepackage{siunitx}

\DeclareMathAlphabet{\mathcal}{OMS}{cmsy}{m}{n}

\input{mysymbol.sty}

\begin{document}

\title{\LARGE \bf Multi-Robot Coverage and Exploration \\ using Spatial Graph Neural Networks}

\author{Ekaterina Tolstaya$^{1}$,  James Paulos$^{2}$, Vijay Kumar$^{1,2}$, Alejandro Ribeiro$^{1}$
\thanks{Supported by ARL Grant DCIST CRA W911NF-17-2-0181, NSF Grant CNS-1521617,
ARO Grant W911NF-13-1-0350, ONR Grants N00014-20-1-2822 and
ONR grant N00014-20-S-B001, and
Qualcomm Research. The first author acknowledges support from the NSF Graduate Research Fellowship.}
\thanks{$^{1}$ Dept. of Electrical and Systems Eng.,
University of Pennsylvania, USA
        {\tt\small eig@seas.upenn.edu}}%
\thanks{$^{2}$Dept. of Mechanical Eng. and Applied Mechanics, University of Pennsylvania, USA}%
}

\maketitle
\thispagestyle{empty}

\begin{abstract}
The multi-robot coverage problem is an essential building block for systems that perform tasks like inspection, exploration, or search and rescue. We discretize the coverage problem to induce a spatial graph of locations and represent robots as nodes in the graph. Then, we train a Graph Neural Network controller that leverages the spatial equivariance of the task to imitate an expert open-loop routing solution.  This approach generalizes well to much larger maps and larger teams that are intractable for the expert. In particular, the model generalizes effectively to a simulation of ten quadrotors and dozens of buildings in an urban setting. We also demonstrate the GNN controller can surpass planning-based approaches in an exploration task. 
\end{abstract}

\input{intro.tex}

\input{problem.tex}
\input{methods.tex}
\input{results.tex}



\bibliographystyle{IEEEtran}
\bibliography{myIEEEabrv,references}

\addtolength{\textheight}{-12cm}   


\end{document}

%% file: intro.tex
\section{Introduction}
Large scale swarms of robots could be deployed to provide on-demand wireless networks \cite{sharma2016uav}, perform rapid environmental mapping \cite{thrun2000real, thrun2005multi}, track targets \cite{schlotfeldt2018anytime}, search after natural disasters \cite{baxter2007multi, jennings1997cooperative}, or enable sensor coverage in communication-denied environments \cite{zhang2005maintaining}. At moderate scales, it may be possible to centralize the entire team's information and control in one agent, but practical deployments of very large teams require distributed execution and scalable algorithms. In particular, global planning-based approaches suffer from an exponential increase in complexity as the number of robots and the environment size increases. This motivates the use of heuristics in general and, as we advocate in this paper, the use of learned heuristics. 

 Graph neural networks (GNNs) have been used to generate heuristic solutions to a variety of multi-robot problems, such as path planning \cite{chen2019learning, battaglia2018relational, joshi2019efficient},  exploration \cite{chen2019self}, and perimeter defense \cite{paulos2019decentralization}. In large teams, we can take advantage of graph equivariance to design abstractions that further speed up learning. In this work, we focus on the problem of coverage, in which a robot team must visit a set of locations in an environment \cite{galceran2013survey}. To leverage recent advances in graph neural networks, we encode the task as a graph: the known map locations and team members are graph nodes, and allowed moves are graph edges. This approach allows us to abstract away the global agent and obstacle locations and to represent all elements of the problem in a single spatial graph with only local connectivity.  Most importantly, the spatial graph representation can describe tasks with complex spatial constraints, unlike past work that applies GNNs to homogeneous inter-robot communication graphs \cite{tolstaya2019learning, khan2019graph}. 

A moderate-size coverage task with dozens of goals and fewer than ten agents can be solved with existing approaches when posed as a vehicle routing problem \cite{toth2002vehicle, ortools}. We collect a dataset of trajectories generated using the centralized expert solution and use behavior cloning to train a graph neural network controller to imitate the expert solution. This learned heuristic can then generalize to previously unseen coverage scenarios with more agents and larger maps. We also show generalization to a scenario with simulated quadrotors that must traverse an environment with thousands of waypoints. 
Furthermore, we apply this approach to an exploration task, which is similar to the coverage task but the graph of waypoints is revealed to the robot team during mission execution. We demonstrate that by imitating an omniscient expert, our learned controller can outperform both a greedy controller and a receding horizon planner. 

For the applications of coverage and exploration, information about distant points of interest may be necessary for computing the next position for each robot. To enable the learned controllers to use distant information, we build graph neural networks with a larger number of graph operation layers, up to 19 layers. The number of graph operation layers determines the distance along which information can travel from one node to another along the edges in the graph, the \emph{receptive field} of the architecture. Other applications of GNNs to robot teams typically use receptive fields of 2 to 4 in conjunction with dense adjacency matrices \cite{chen2019self, khan2019graph, tolstaya2019learning, sykora2020multi}, with the exception of the shortest path demonstration in \cite{battaglia2018relational} with a receptive field of 10. A mean aggregation operation helps stabilize training of GNNs with larger receptive fields. Furthermore, we use a sparse representation of the local connectivity of the graph that allows our approach to scale to larger maps and teams than \cite{sykora2020multi}.
Our contributions can be summarized as follows:
\begin{enumerate}
\item An approach to encoding continuous space multi-robot coverage and exploration problem data as a discrete spatial graph in which robots and points of interest are nodes and allowed moves are edges.
\item A training methodology using behavior cloning with an optimization based VRP expert solution for the coverage problem, as well as an extension to exploration with partially observable states.
\item  A graph neural network architecture which explicitly respects equivariance in the task structure in order to achieve zero shot generalization to large maps and large teams for which an expert demonstration is intractable.
\end{enumerate}

\begin{figure*}[t]
\centering
\vspace{0.15cm}
\begin{subfigure}[b]{.32\linewidth}
\includegraphics[width=1.0\textwidth]{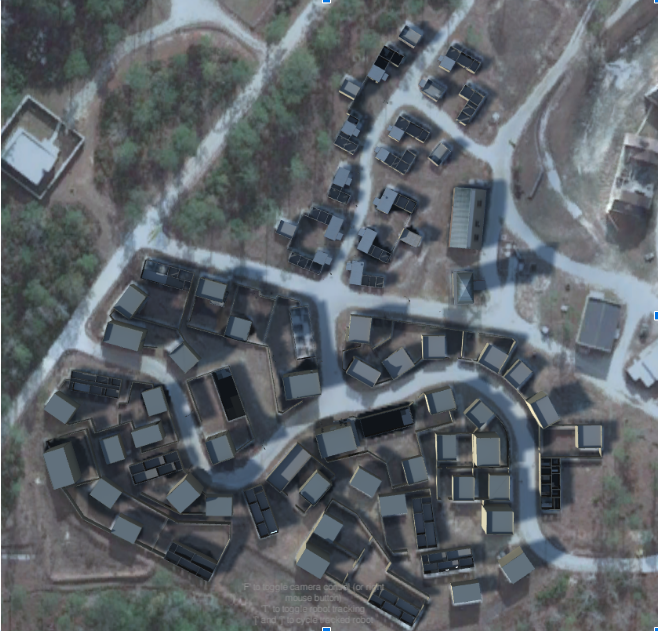}
\caption{A city simulated in Unity. }\label{fig:lejeune}
\end{subfigure}
\begin{subfigure}[b]{.32\linewidth}
\includegraphics[width=1.0\textwidth]{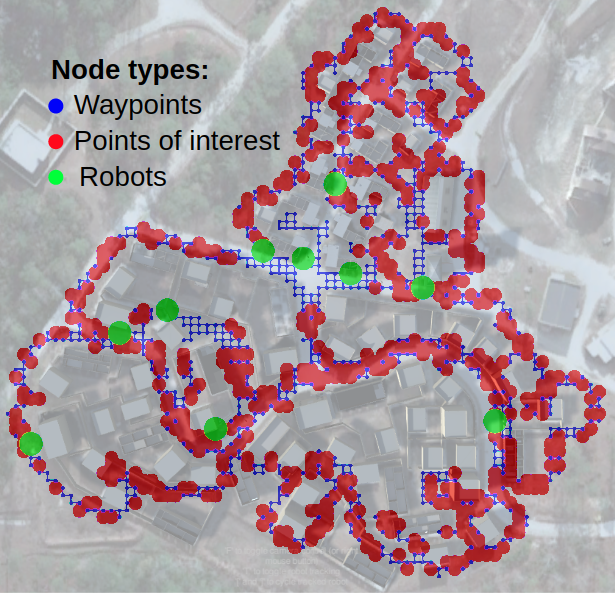}
\caption{The graph representation of the  task.}\label{fig:overlay}
\end{subfigure}
\begin{subfigure}[b]{.325\linewidth}
\includegraphics[width=1.0\textwidth]{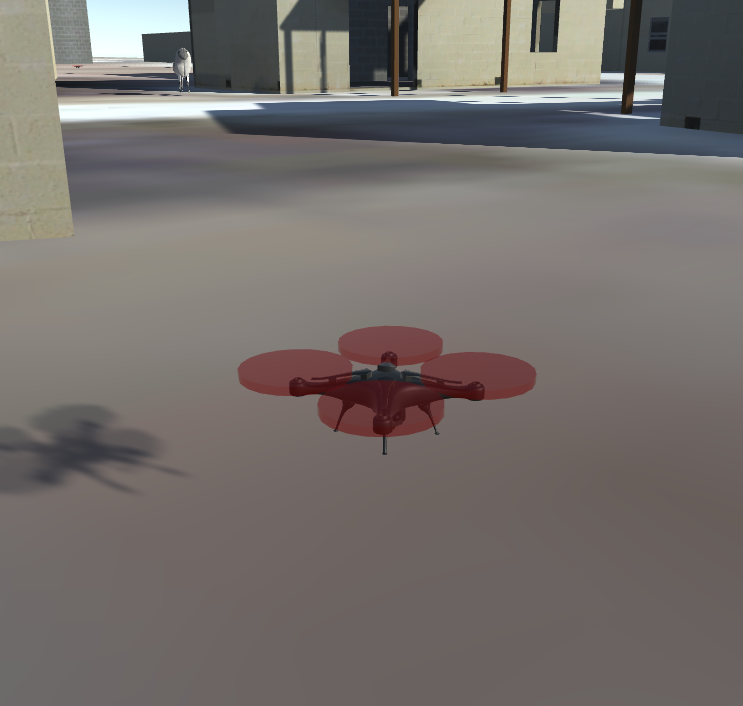}
\caption{A team of 10 such quadrotors were used. }\label{fig:quad}
\end{subfigure}
\caption{The trained models were tested on a team of robots simulated in Unity and controlled by waypoint commands issued through a Robot Operating System interface. The trained model allows the robots to divide and conquer to visit the points of interest more efficiently than a greedy model. We visualize this experiment in a  provided along with this work: \url{https://youtu.be/MiYSeENTyoA} }\label{fig:lejeune_coverage}
\end{figure*}

%% file: problem.tex
\section{Multi-Robot Routing for the Coverage Problem}

We consider the time-constrained coverage path planning problem in which a team of robots must maximize the visited region of interest within a given time limit \cite{galceran2013survey}. The problem of searching over the set of all feasible trajectories is intractable, so we finely discretize the region of interest, motivated by the success of motion planning in lattices \cite{mcnaughton2011motion}. In practice, a coarse map of a region may be provided via satellite imagery. To generate the lattice representation, we initialize a grid of points with a spacing of 5 meters, and then remove all points that are within an obstacle in the mission environment pictured in Fig. \ref{fig:lejeune_coverage}, and then add connections between adjacent nodes in free space.

We denote the set of robots as $\mathcal{R}$,  waypoints as $\mathcal{W}$, and unvisited waypoints of interest denoted as $\mathcal{X}$. The set of unvisited waypoints of interest is a subset of all waypoints, $\mathcal{X} \subseteq \mathcal{W}$. Next, we define the parameters that describe the map of the environment: $p^j$ is the location of waypoint $j$, and $\mathcal{N}_j$ is the set of neighboring waypoints to waypoint $j$. Then, we define the mission: $x^j_t \in \lbrace 0, 1 \rbrace$ is an indicator of whether the waypoint $j$ is of interest at time $t$, with $x_t^j = 1$ if $j \in \mathcal{X}$. $T$ denotes the time budget for the mission.  The location of robot $i$ at time $t$ is $q_t^i$ and $\mathbb{1}_{q_{t}^i = p^j}$ indicates whether robot $i$ is currently at waypoint $j$.
%
The multi-robot coverage problem can now be formulated as:
\begin{align}
\max_{\lbrace q_{t}^i \rbrace_{i \in \mathcal{R}}} \;\; & \sum_{t=0}^T \sum_{j \in \mathcal{W}} \sum_{i \in \mathcal{R}} x^j_{t}  \mathbb{1}_{q_{t}^i = p^j} \\
\textrm{s.t.} \;\;
& x^j_t = x^j_{0} \prod_{i \in \mathcal{R}} \prod_{s=0}^{t-1} (1 - \mathbb{1}_{q_{s}^i = p^j}), \quad \forall j \in \mathcal{W}, t \leq T  \nonumber \\ 
&   q_{t-1}^i \!\!=\! p^j , q_{t}^i \!=\! p^k \Rightarrow k \in \mathcal{N}_j,  \;\; \forall i \in \mathcal{R}, \forall j,k \in \mathcal{W}, t \leq T. \nonumber
\end{align}
%
Alternate formulations exist that can pose the coverage problem as a Vehicle Routing Problem (VRP) to be solved as a mixed-integer program \cite{dantzig1954solution}.
For a given problem instance, we can use existing routing solvers such as \cite{ortools} to generate open loop solutions. 
Other approaches to the coverage problem typically pre-compute the roles and trajectories of all team members for the duration of the mission \cite{galceran2013survey}. 

In contrast, our goal is to develop a closed-loop controller that computes only the next action for each robot based on the current state of the system. This approach generalizes easily to systems with real dynamics in which agents may not instantaneously transition between desired waypoints, such as the team of simulated robots in Fig. \ref{fig:lejeune_coverage}. It also enables generalization to dynamic graphs, permitting us to extend the coverage problem to an exploration problem in which new waypoints are discovered online during execution.
We seek to learn a closed loop controller, $\pi$, that maximizes the observed waypoints in expectation over the set of initial states and maps:
%
\begin{align}
\label{eq:closed_loop}
\max_\pi \;\;  & \mathbb{E}_{q_0^i, p^j, x_0^j} \;\; \sum_{t=0}^{T} \sum_{j \in \mathcal{W}} \sum_{i \in \mathcal{R}} x^j_{t} \; \mathbb{1}_{q_{t}^i = p^j} \\
\textrm{s.t.} \;\;  
& x^j_t = x^j_{0} \prod_{i \in \mathcal{R}} \prod_{s=0}^{t-1} (1 - \mathbb{1}_{q_{s}^i = p^j}), \quad \forall j \in \mathcal{W}, t \leq T  \nonumber \\
&  q_{t-1}^i \!=\! p^j , q_{t}^i \!=\! p^k \Rightarrow k \in \mathcal{N}_j, \;\;\; \forall i \in \mathcal{R}, \forall j,k \in \mathcal{W}, t \leq T \nonumber \\
& \lbrace q_{t}^i \rbrace_{i\in\mathcal{R}}= \pi\left(\lbrace q_{t-1}^i\rbrace_{i\in\mathcal{R}}, \;\; \lbrace x_{t-1}^j\rbrace_{j\in\mathcal{W}}, \;\;\lbrace p^j\rbrace_{j\in\mathcal{W}}\right).  \nonumber 
\end{align}
%

%% file: methods.tex
\section{Methods}

To solve the problem of multi-robot coverage, we develop a parametrization of the problem as a heterogeneous graph that can be input to a Graph Neural Network to  be trained via supervised learning. 

\subsection{Graph Representations for Coverage}

We develop a parametrization of the current system state in which all entities, including robots and waypoints, are part of a single computation graph. Each robot in the team is considered as a node in the graph and robots are allowed to move from waypoint to waypoint in discrete steps.  Due to robots' movement, the graph topology changes during execution.
Using the lattice representation, we can abstract away the global positions of the robots, and maintain only relative information of nearby waypoints and robots. Unlike \cite{chen2019self}, we do not need to provide the relative distances of all waypoints to all agents. Furthermore, the action space of each robot is now discrete: the robot chooses one nearby waypoint to move to.

There are two types of edges: 1) map edges between waypoints to indicate free space and 2) action edges between robots and waypoints that indicate a robot's capability to move to nearby locations. Both types of edges are undirected.
The waypoint connectivity is defined by the lattice induced over the given map of obstacles and free space. A robot is connected to the same waypoints as the waypoint at its current location, if $q_i = p_j$, then $\mathcal{N}_i = \mathcal{N}_j$. 

\begin{figure}[t]
\centering
\includegraphics[width=\textwidth]{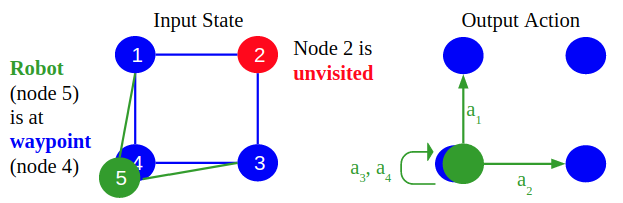}
\caption{Robots and waypoints comprise the nodes in the graph, with the edges between them indicating the ability of robots to move to new locations. 
}\label{fig:nodes_edges_cartoon}
\end{figure}

The feature vector for each node $\bbv_i$ of index $i$ indicates the type of this node: 
\begin{equation} \label{eq:node_features}
    \bbv_i = \lbrace \mathbb{1}_{i \in \mathcal{R}}, \mathbb{1}_{i \in \mathcal{W}}, \mathbb{1}_{i \in \mathcal{X}}   \rbrace,
\end{equation}
where $\mathbb{1}$ is an indicator function. 

We define $\bbe_k$ as an edge feature vector for the directed edge of index $k$, with a sender node $s_k$ and a receiver node $r_k$.
For this task, we define $\bbe_{k}$ to be the distance between the positions of nodes $s_k$ and $r_k$:
\begin{equation} \label{eq:edge_features}
    \bbe_{k} = \lVert p_{s_k} - p_{r_k} \rVert.
\end{equation}


The set of all edge features is $E = \lbrace \bbe_{k} \rbrace$, and the set of vertex features is $V = \lbrace \bbv_i \rbrace$, and we denote the graph that represents the state of the entire system as $\mathcal{G} = \{E,V\}$. At a given time $t$, the state of the multi-robot task is described by $\mathcal{G}_t$.




\subsection{Graph Representations for Exploration}

We view the exploration problem as the problem of coverage on a growing graph. 
Waypoint nodes are added to the graph when they are observed by a range sensor with range $S$: if $ \lVert p^i_t - q^j_t \rVert \leq S$, then $\mathcal{W}_{t+1} = \mathcal{W}_t \cup \lbrace p_i \rbrace$, with the set of waypoints growing over time.
Exploration introduces the possibility that an observed waypoint may or may not have adjacent waypoints that are currently unexplored. We call these frontier nodes and add an indicator feature to indicate whether a waypoint is part of the set of frontier nodes, $\mathcal{F}$: 

\begin{equation} \label{eq:node_features_explore}
    \bbv_i = \lbrack \mathbb{1}_{i \in \mathcal{R}}, \mathbb{1}_{i \in \mathcal{W}}, \mathbb{1}_{i \in \mathcal{X}} , \mathbb{1}_{i \in \mathcal{F}}  \rbrack.
\end{equation}

\subsection{Aggregation Graph Neural Networks}

Graph Neural Networks are an increasingly popular tool for exploiting the known structure of any relational system \cite{battaglia2018relational}.  In graph convolutional networks, the graph convolution operation is defined using learnable coefficients that multiply powers of the adjacency matrix times the graph signal \cite{kipf17-classifgcnn, gama2018convolutional}. We extend this architecture by incorporating non-linear graph convolution operations.

The building block of a GNN is the Graph Network Block. 
Given a graph signal, $\mathcal{G} =\big\lbrace  \{ {\bbe}_{k} \}, \{{\bbv}_{i}\} \big\rbrace$, one application of the GN block transforms these features, $\mathcal{G}' = \big\lbrace \{ {\bbe}_{k}' \}, \{ {\bbv}_{i}'\}\big\rbrace$:
\begin{align} \label{eq:gnn_block}
    \bbe_k' = \phi^e(\bbe_k, \bbv_{r_k}, \bbv_{s_k}) , \;\;\;
    \bbv_i'  = \phi^v(\bar{\bbe}_i', \bbv_i) , \;\;\;
    \bar{\bbe}_i'  = \rho^{e \rightarrow v}(E_i').
\end{align}

$GN(\cdot)$ is a function of the graph signal $\mathcal{G}$, described by the application of $\phi^e$, $ \rho^{e \rightarrow v}$ and $\phi^v$ in that order to produce the transformed graph signal $\mathcal{G}'$, with the same connectivity but new features on the edges and nodes. 

The aggregation operation $\rho^{e \rightarrow v}$ takes the set of transformed incident edge features $E'_i=\lbrace \bbe'_k \rbrace_{r_k = i}$ at node $i$ and generates the fixed-size latent vector $\bar{\bbe}_i'$. Aggregations must satisfy a permutation invariance property since there is no fundamental ordering of edges in a graph. Also, this function must be able to handle graphs of varying degree, so the mean aggregation is particularly suitable to normalizing the output by the number of input edges \cite{tolstaya2019learning}:
\begin{equation}
    \rho^{e \rightarrow v}(E_i') := \frac{1}{  |E'_i|} \sum_{\bbe_k' \in E'_i} \bbe_k'.
\end{equation}
The mean aggregation operation is especially helpful for improving the stability of GNNs with large receptive fields. 

Next, we describe two variants of the Aggregation GNN architecture that build upon \cite{gama19aggGNN}. 
The linear Aggregation GNN architecture uses the following parametrization:
\begin{align} \label{eq:linear_gnn}
    \phi_L^e(\bbe_k, \bbv_{r_k}, \bbv_{s_k}) := \bbv_{s_k} , \;\;\;\;\;
    \phi_L^v(\bar{\bbe}_i', \bbv_i)  := \bar{\bbe}_i', 
\end{align}
while the non-linear Aggregation GNN uses learnable non-linear functions to update node and edge features:
\begin{align} \label{eq:nonlinear_gnn}
    \phi_N^e(\bbe_k, \bbv_{r_k}, \bbv_{s_k}) &:=\text{NN}_e( \lbrack \bbe_k, \bbv_{r_k}, \bbv_{s_k}\rbrack ), \nonumber \\
    \phi_N^v(\bar{\bbe}_i', \bbv_i)  &:= \text{NN}_v( \lbrack \bar{\bbe}_i', \bbv_i \rbrack), 
\end{align}
where $\text{NN}_e$ and $\text{NN}_v$ are 3 layer MLPs with 16 hidden units. 
Note that the linear Aggregation GNN in \eqref{eq:linear_gnn} cannot use the input edge features, such as those defined in \eqref{eq:edge_features}, unlike the non-linear GNN defined in \eqref{eq:nonlinear_gnn}.


\begin{figure}[h]
\centering
\includegraphics[width=\textwidth]{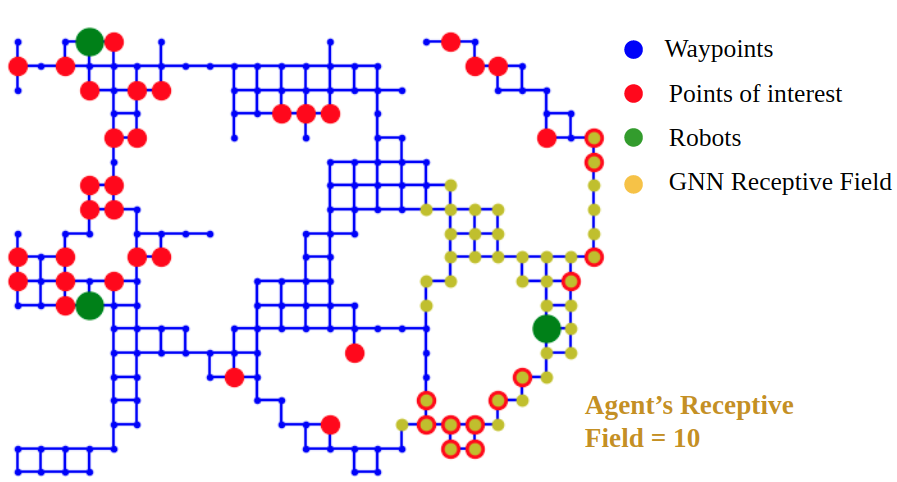}
%
\caption{The receptive field (K=10) of a Graph Neural Network. 
}\label{fig:rf_cartoon}
\end{figure}

\subsection{Policy Architecture}

While a Graph Network Block can be used to compose a variety of architectures, for this work, we develop a variant of the Aggregation GNN in which the output of every GN stage is concatenated, and finally processed by a linear output transform \cite{gama19aggGNN}:

\begin{align} \label{eq:enc_proc_dec}
\mathcal{G}' = f_{\text{out}} \Big( \big\lbrack f_{\text{dec}} (f_{\text{enc}}(\mathcal{G})),  f_{\text{dec}} (GN(f_{\text{enc}}(\mathcal{G}))), \;\;\; \\ \nonumber  
\quad  f_{\text{dec}} (GN(GN(f_{\text{enc}}(\mathcal{G})))),  \;\; ...  \;\; \big\rbrack \Big).
\end{align}

The addition of the encoder $f_{\text{enc}}$ and decoder $f_{\text{dec}}$ layers was inspired by the Encode-Process-Decode architecture presented in \cite{battaglia2018relational}. The number of $GN$ operations is a hyperparameter and determines the \textit{receptive field} ($K$), of the architecture, or how far information can travel along edges in the graph. A GNN with a receptive field of zero can be compared to the Deep Set architecture that neglects any relational data \cite{NIPS2017_f22e4747}.

In our policy architecture, $f_{\text{enc}}$, $f_{\text{dec}}$ are 3 layer MLPs with 16 hidden units, and a ReLU activation only after the first two layers. 	
$f_{\text{out}}$ is a linear function that reduces the high-dimensional latent space vectors output on the edges to the the logits of a Boltzmann distribution. We sample from the edges that connect each robot node to neighboring waypoint nodes to determine the edge that each robot will next travel.
This is in contrast to \cite{tolstaya2019learning} where there was a hand-engineered feature extractor. 


Fig. \ref{fig:rf_cartoon} visualizes a segment of a typical training scenario. Three robots shown in green must visit red points of interest by traveling along the waypoints in free space indicated by blue nodes and edges. We visualize the receptive field of a GNN with $K=10$ indicating the 10-hop information available to the robot on the right. A larger receptive field allows each agent to use information about more distant regions of the map to compute the controller. 

\subsection{Baseline Controllers}

The learned policies are compared to three types of controllers: 1) an expert open loop VRP solution, 2) a receding horizon VRP-based controller, and 3) a greedy controller. 
We use Google's OR-Tools library \cite{ortools} to provide optimization-based expert solutions to the VRP. 
This expert plans once for the full mission length of $T$ and then this trajectory is executed in an open loop fashion. The expert assumes global knowledge of the map and may be intractable in larger or dynamic graphs. The training data was generated with this open-loop approach. We also devise a receding horizon controller that plans for trajectories of $\hat{T} < T$, and then executes the first step of this trajectory, and then re-plans at the next time step. The Expert controller baselines use receding horizon control in Figs. \ref{fig:coverage}, \ref{fig:explore}, \ref{fig:coverage_rl}.
Finally, a greedy controller that routes each robot to the nearest unvisited point of interest is a great heuristic in many scenarios, so we include it as a practical lower bound. The greedy controller can be implemented with a finite receptive field using only a K-step distance matrix between nodes in the graph and we provide this benchmark in Figs. \ref{fig:coverage}, \ref{fig:explore}, \ref{fig:coverage_rl}. A limited-horizon greedy controller may be more practical in larger graphs for which computing a full distance matrix is expensive.

\subsection{Imitation Learning from Expert Demonstrations}

In behavior cloning, our goal is to use stochastic gradient descent to minimize the difference between the expert's action and the policy's output, where $\ccalL$ is a cross-entropy loss, since the action space is discrete:
\begin{equation} \label{eq:empirical_loss}
    \pi^{\ast} = \argmin_{\pi} \sum_{(G_t,\bbu_t) \in \ccalD}
                   \ccalL\Big( \pi(G_t), \bbu_{t} \Big).
\end{equation}
To use behavior cloning to train the graph neural network policy, we require a dataset of expert trajectories. We collect a dataset of 2000 expert trajectories of length $T=50$ in randomly generated graphs, $\ccalD = \lbrace (G_t, \bbu_t) \rbrace_{t=1,\ldots,50}$. The graphs are generated by sampling regions of 228 waypoints on average from the graph shown in \ref{fig:lejeune_coverage}. The performance of the learned controller is tested on graphs generated from the same distribution over trajectories of length $T=50$. 
The models were trained for 200 epochs with a batch size of 32. Adam optimizer was used with an initial learning rate of 0.001 which was decayed by a factor of 0.95 for every 200 batches.

For the exploration task, we use the expert controller that uses the full graph to generate the trajectory, but only the local state observations are stored in the dataset, so the robot is learning to predict what the omniscient centralized controller would do based only on partial observations of currently explored nodes and frontiers.

\subsection{Implementation Details} \label{sec:implementation}
We use DeepMind's Graph Nets library and a variant of the Encode-Process-Decode architecture for the graph neural network policy \cite{battaglia2018relational}. 

We implement a local collision avoidance strategy as part of the task specification. A robot is allowed to move to a new waypoint if no other robot will be occupying this waypoint during the next time step. Conflicts are resolved by giving priority to the robot with the smaller index. If agents moved to the same waypoint, they are likely to continue to travel together since the policy cannot disambiguate the agents due to the graph parametrization. To avoid this redundancy, we eliminate the ability of robots to move to the same waypoint. In a real decentralized multi-robot team, collision avoidance based on on-board sensing may also be necessary. 

Due to the known maximum degree of the map graph, we can fix the number of neighboring waypoints considered by the policy to be up to 4, so that existing infrastructure for stochastic policies with fixed-size action spaces can be used \cite{stable-baselines}. The output of the learned policy are weights for up to 4 edges from waypoints to robots, as shown in Fig. \ref{fig:nodes_edges_cartoon}.

Finally, an open-source implementation of the learning architecture can be found here: \url{https://github.com/katetolstaya/graph_rl} and the applications here: \url{https://github.com/katetolstaya/gym-flock}.

%% file: results.tex
\section{Results}

First, we highlight the impact of the Aggregation GNN's receptive field on its performance on the exploration and coverage tasks. Next, we examine how the GNN can generalize to larger graphs and team sizes. Finally, we validate the use of the GNN controllers in a high-fidelity simulator. 


\subsection{Locality}

\begin{figure}
    \centering
    \includegraphics[width=\textwidth]{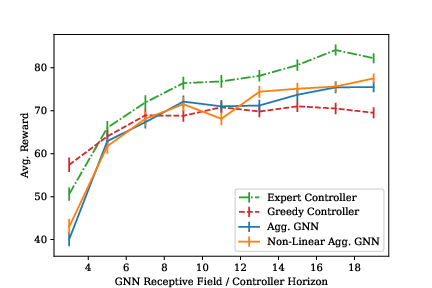}
    \caption{GNNs with larger receptive fields achieve higher rewards on the coverage task. Mean reward over 100 episodes with standard error is shown.  
    }
    \label{fig:coverage}
\end{figure}

On the coverage task, the learned controllers reliably outperform a greedy controller, but fall short of the receding-horizon expert in Fig. \ref{fig:coverage}. For the GNNs, we see a sharp improvement in performance with increasing receptive field. The linear and non-linear variants of the aggregation GNNs perform comparably, with the non-linear GNN performing slightly better.
With an increasing horizon, the greedy controller rapidly reaches its asymptotic mean reward of 70.1 with a standard error of the mean (SEM) of 1.32. The open-loop expert obtains a mean reward of 91.0 with a SEM of 0.87. In this experiment, the expert controller is optimal and obtains the maximum rewards possible and provides the upper bound on the performance of the GNN. 

\begin{figure}
    \centering
    \includegraphics[width=\textwidth]{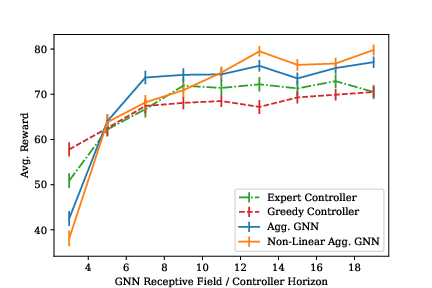}
    \caption{GNNs surpass the expert controller on the exploration task. Mean reward over 100 episodes with standard error is shown. 
    }
    \label{fig:explore}
\end{figure}

On the exploration task, the learned controllers outperform greedy and the planner-based controllers as we can see in Fig. \ref{fig:explore}.
Existing planning-based solutions rely on ad-hoc heuristics to  weigh the importance of a waypoint at the frontier versus other waypoints of interest, and we show that a learned solution can improve over a receding horizon expert controller.  We also see a significant improvement in the mean reward obtained by the GNNs as their receptive field increases. The non-linear GNN is again slightly better than the linear GNN. 

\begin{figure}
    \centering
    \includegraphics[width=\textwidth]{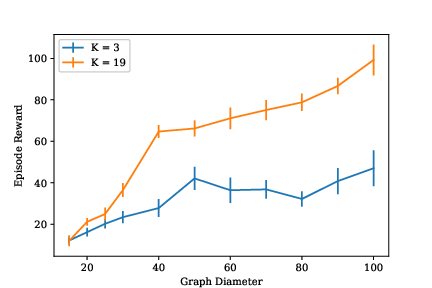}
    \caption{Effect of receptive field of non-linear GNNs in graphs of varying diameters, as measured by the mean reward over 20 episodes and standard error. 
    }
    \label{fig:rf_result}
\end{figure}

We further analyze the effect of a model's receptive field on its performance in varying size graphs. A finite receptive field decomposes the problem into local neighborhoods for each robot, producing a decentralized solution. The robot only uses a fixed-hop neighborhood of the graph for computing the action as seen in Fig. \ref{fig:rf_cartoon}. The graph diameter is the max distance between any two nodes in a graph. A model with a smaller receptive field performs worse than a model with a larger receptive field, especially in graphs of larger diameter as demonstrated by Fig. \ref{fig:rf_result}.
As the graph size increases, there are more points of interest that must be visited. The larger receptive field controller is able to route the agents to these waypoints, while the controller with the smaller receptive field ($K=3$) is unable to compute a high-reward path.

While the expert controller outperforms the learned controllers in Fig. \ref{fig:coverage}, the VRP-based controller requires two orders of magnitude more time to compute a control action, as seen in Table \ref{runtime_table}. 
The neural networks used a GTX 1080 GPU and the benchmarks used one core of an Intel Core i9-9900K processor. The expert and greedy controllers with receptive field of $\infty$ use the full adjacency matrix and, in particular, the expert plans one open-loop trajectory of length $T=50$. We observe that the linear GNN has a faster controller computation time, and controllers with larger receptive fields also need more time.

\begin{table}[!tbp]
\label{tab:runtime}
\centering
\begin{tabular}{lccc}
\toprule
& & Receptive Field  & \tabularnewline
Policy & K=9  & K=19 & $\infty$   \tabularnewline
\midrule
Expert   & \num{13500}   & \num{23500} & \num{2330}
 \tabularnewline
 GN-MLP   & \num{176}  & \num{277} & -
 \tabularnewline
 GN-Linear   & \num{133}   & \num{171} & -
 \tabularnewline
  Greedy   & \num{86.3}  & \num{142} & \num{297}
 \tabularnewline
\bottomrule
\vspace{-0.2cm}
\end{tabular}
\caption{
\small The average controller computation time per episode in milliseconds, tested over 100 episodes of the coverage task.  The expert controller (Receptive field = $\infty$) plans one open-loop trajectory for the task given the full adjacency matrix. 
}
\label{runtime_table}
\end{table}

\subsection{Transference}

\begin{figure}
    \centering
    \includegraphics[width=\textwidth]{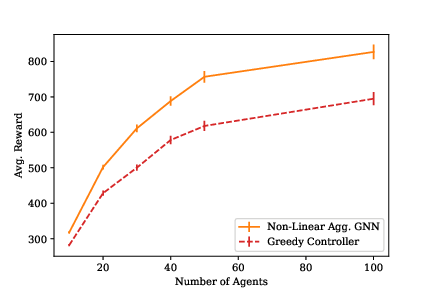}
    \caption{Generalization to a coverage task with 5659 waypoints. We plot the average reward over 100 episodes with standard error. The GNN was trained with 4 agents and tested on teams of up to 100.
    }
    \label{fig:coverage_generalize}
\end{figure}

\begin{figure}
    \centering
    \includegraphics[width=\textwidth]{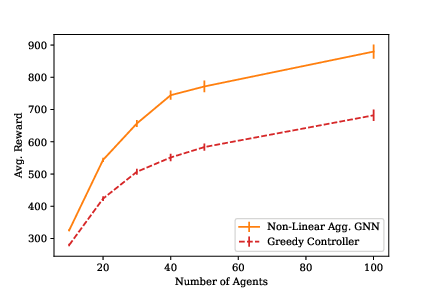}
    \caption{Generalization to an exploration task with 5659 waypoints. We plot the average reward over 100 episodes with standard error. The GNN was tested on teams of up to 100. 
    }
    \label{fig:explore_generalize}
        
\end{figure}

The trained GNN models effectively generalize to larger robot team and map sizes than can be solved by conventional VRP solutions. 
The models were first trained on 4 agents and 228 waypoints on average. Then, the models were tested on a map size of 5659 waypoints with a graph diameter of 205. The team size varied from 10 to 100 agents. For both the coverage and exploration generalization experiments, the map and team sizes made the centralized expert solution intractable.
In Fig. \ref{fig:coverage_generalize}, we observe that the learned solution consistently outperforms the greedy controller on the coverage task.  In Fig. \ref{fig:explore_generalize}, this difference is even bigger for the exploration task. We hypothesize that this is because the learned policy learns to weigh the frontier nodes more than other unexplored nodes. 

\subsection{Reinforcement Learning}

\begin{figure}
    \centering
    \includegraphics[width=\textwidth]{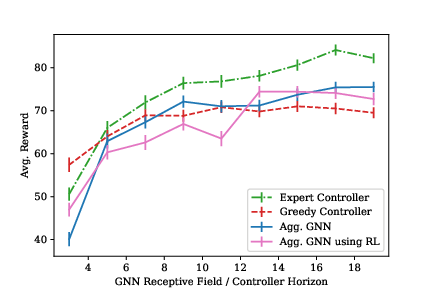}
    \caption{Comparison of linear GNNs trained with imitation learning and reinforcement learning. Mean reward over 100 episodes with standard error. 
    }
    \label{fig:coverage_rl}
\end{figure}

Reinforcement Learning (RL) is also an effective tool for training Aggregation GNN controllers and achieves comparable performance to imitation learning in Fig.  \ref{fig:coverage_rl}. RL is especially important in tasks that do not have suitable expert controllers for generation of training data. We use the Proximal Policy Optimization algorithm \cite{schulman2017proximal} and parametrize both the policy and value functions as Aggregation GNNs. The policy architecture is the same as for the imitation learning experiments, while the value function sums over the values output for the robot nodes. The GNN captures the known structure of the problem for the value and policy functions.
The reward is the summation in the objective of \eqref{eq:closed_loop}, with no reward shaping required.
The model was trained using $1 \times 10^6$ observations using an open source implementation of PPO from \cite{stable-baselines} using Adam optimizer with a step size of $1 \times 10^{-4}$ and a batch size of 40.
The controller trained using RL reached the performance of the imitation-trained controller for higher receptive fields. We observed a higher variance in the performance of the controllers trained with RL, so additional tuning of this system may be necessary. 

\subsection{Dynamics}

Despite training on small-scale teams with instantaneous discrete state transitions, the GNN is effective for control in a coverage mission with ten quadrotors in a large simulated environment, pictured in Fig.  \ref{fig:lejeune_coverage}. 
In 400 seconds of mission time, the team of 10 robots visited 490 points of interest using the greedy controller, as compared to visiting 610 points of interest using the non-linear GNN with a receptive field of 19. 
The use of the lattice representation with discrete states allows the model trained on an ideal discrete task to generalize zero-shot to a high-fidelity simulator. We discretize the robot positions provided by the simulator by clipping them to each robot's nearest waypoint.
The spatial aggregation operations are invariant to the time-scale of the task, so we can execute the GNN controller with the non-ideal dynamics of a quadrotor. 
The local collision avoidance strategy described in Section \ref{sec:implementation} was effective for this scenario because the implementation of the control policy and the evaluation of the GNN was centralized. Furthermore, each robot was assigned a different altitude. In more complex environments or for tasks specified on a 3D lattice, a policy that uses on-board sensing may be required.

\section{Concluding Remarks}

We develop a scalable GNN architecture for multi-robot coverage and exploration tasks. The approach surpasses existing decentralized heuristics and also scales well to from team sizes of 4 to teams of up to 100 agents. We also demonstrate that this architecture can be trained via reinforcement learning.  As a bridge to deploying this approach to physical robot teams, we demonstrate generalization to a simulated robot team subject to dynamics in a dense urban environment. 

To deploy the GNN in a real distributed team, we would need to address challenges such as asynchronous or intermittent communication. One possible solution could be the evaluation of the GNN on the contents of a robot's local buffer containing the estimated state of the system, and allowing intermittent communication among robots to update each other about the current positions of other robots, points of interest, and, for the exploration task, the growing map of waypoints. One approach to enable data distribution in mobile robot teams was explored in \cite{tolstaya2021learning}.






